%% file: acl_latex.tex
\pdfoutput=1

\documentclass[11pt]{article}

\usepackage[]{acl}

\usepackage{times}
\usepackage{adjustbox}
\usepackage{multirow}
\usepackage{latexsym}
\usepackage{amsmath}
\usepackage{amssymb}
\usepackage{makecell}
\usepackage{amssymb}

\usepackage{mathabx}
\usepackage[T1]{fontenc}

\usepackage[utf8]{inputenc}

\usepackage{listings}
\usepackage{xcolor}

\newtheorem{definition}{Definition}
\graphicspath{{./images/}} 
\usepackage{microtype}

\title{Supervised Gradual Machine Learning For Aspect Category Detection}

\author{ Murtadha Ahmed \quad 
	Qun Chen  \\
	Northwestern Polytechnical University, Xi'an, China \\
	{\tt \{a.murtadha@mail.,chenbenben@\}nwpu.edu.cn}
}

\begin{document}
\maketitle
\begin{abstract}
	

	Aspect Category Detection (ACD) aims to identify implicit and explicit aspects in a given review sentence. The state-of-the-art approaches for ACD use Deep Neural Networks (DNNs) to address the problem as a multi-label classification task. However, learning category-specific representations heavily rely on the amount of labeled examples, which may not readily available in real-world scenarios.
In this paper, we propose a novel approach to tackle the ACD task by combining DNNs with Gradual Machine Learning (GML) in a supervised setting. we aim to leverage the strength of DNN in semantic relation modeling, which can facilitate effective knowledge transfer between labeled and unlabeled instances during the gradual inference of GML. To achieve this, we first analyze the learned latent space of the DNN to model the relations, i.e., similar or opposite, between instances. We then represent these relations as binary features in a factor graph to efficiently convey knowledge. Finally, we conduct a comparative study of our proposed solution on real benchmark datasets and demonstrate that the GML approach, in collaboration with DNNs for feature extraction, consistently outperforms pure DNN solutions.
\end{abstract}

\input{introduction}
\input{related_work}
\input{background}

\input{approach}

\input{experiment}

\section{Conclusion}
In this paper, we have proposed a novel supervised GML approach for the task of ACD. It first leverages the strength of deep neural networks to construct semantic relations between instances of the same category. Then, it models these relations as binary factors to enable effective knowledge conveyance in a factor graph.  Our extensive experiments have shown that the proposed approach consistently achieves the state-of-the-art performance. For future work, it is interesting to note that the GML approach in collaboration with DNN for feature extraction is potentially applicable to other classification tasks; technical solutions however need further investigation. 

\bibliographystyle{acl_natbib}
\bibliography{acl_references}

\end{document}

%% file: introduction.tex
\section{Introduction}

Aspect Category Detection (ACD) is a crucial subtask of Aspect-Based Sentiment Analysis (ABSA) \cite{liu2012sentiment,wang2016attention,pontiki2014semeval,pontiki2016semeval,AhmedCWL19}. ACD aims to identify the categories explicitly or implicitly mentioned in a given review. For example, consider the running example illustrated in Table \ref{tab:running_example}, where the sentence $s_1$ describes a restaurant in terms of two aspects, namely, \textit{food} and \textit{price}. In this case, the primary goal of ACD is to detect these two aspect categories. 
ACD has received significant research attention due to its crucial role in extracting detailed opinions that are expressed towards specific aspects of an entity \cite{saeidi2016sentihood,jiang-etal-2019-challenge,BERT-ASC}. By detecting and analyzing these aspects, businesses can gain valuable insights into customer preferences, needs, and opinions, which can enable them to make informed decisions about product development, marketing strategies, and customer engagement. Consequently, this can lead to higher levels of customer satisfaction and loyalty \cite{hu2004mining}.


Various methods have been proposed to address ACD, which can be broadly classified into three categories: rule-based systems \cite{schouten2017supervised}; machine learning algorithms \cite{castellucci2014unitor,kiritchenko2014nrc}; and Deep Neural Networks (DNNs) \cite{sun2019utilizing,wu2020context}. Rule-based systems rely on a set of pre-defined rules based on domain-specific knowledge and linguistic patterns to identify aspect categories. On the other hand, machine learning algorithms use statistical models to learn from labeled data and identify aspect categories based on the features of the text. The state-of-the-art solutions for ACD have been built using DNNs \cite{sun2019utilizing,wu2020context}. Recently, ACD has experienced a considerable shift towards fine-tuning pre-trained language models (PLMs), e.g., BERT \cite{devlin2018bert}, to jointly address aspect detection and sentiment. However, ACD remains a very challenging task due to the implicit nature of aspect descriptions in sentences. For instance, in SemEval 2014 Task 4 \cite{pontiki2014semeval}, 73.8\% of sentences describe implicit aspects, which makes it difficult to perform accurately ACD without a substantial amount of labeled examples that can be both costly and time-intensive to procure in real-world settings.




\begin{table*}
		
		\centering

		\begin{tabular}{l|l|l}
			\hline
			$s_i$& \makecell[c]{Sentence Review}&Categories  \\
			\hline	
			
			$s_1$& \textbf{Food} was very  \textbf{expensive} for what you get.& Food, Price\\
			$s_2$&There was only one \textbf{waiter} for the whole restaurant. & Service\\
			
			$s_3$&\textbf{Waiters} are very \textbf{friendly} and the \textbf{pasta} is out of this world. &Service, Food\\
			\hline
		\end{tabular}
			\caption{A running example of ACD with bolded words representing category representatives. }
	\label{tab:running_example}
	
\end{table*}

\citet{wang2019gradual,ahmed-etal-2021-dnn} have demonstrated that unsupervised Gradual Machine Learning (GML) can effectively perform Aspect-Based Sentiment Detection (ABSD). It is noteworthy that ABSD operates under the assumption that the aspects are predefined, and the task is to detect the corresponding sentiment.
Generally, GML begins with automatically labeled easy instances and gradually reasons about the labels of more challenging instances through iterative knowledge conveyance  \cite{hou2019gradual,hou2020gradual}. As a non-i.i.d learning paradigm, GML utilizes shared features between labeled and unlabeled instances for knowledge conveyance. It is noteworthy that the current GML-based solutions for ABSD operate in an unsupervised setting and does not rely on labeled examples for training.
Although they can achieve competitive performance compared to traditional supervised DNN-based models, they still have limitations in terms of accuracy and knowledge conveyance. These limitations pose two key challenges when applying them to the ACD task. First, the current feature extraction method relies on sentiment lexicons and shift expressions (such as but or however). While these feature are useful indicators of sentiment, they may not necessarily indicate the aspect being discussed, which can limit accuracy. Second, the implicit nature of aspects makes it difficult to design reliable rules that accurately capture their representatives.

In this paper, we present a novel supervised approach for the ACD task that is based on the GML paradigm that leverages labeled examples to enable more effective gradual learning. Specifically, we combine strength of DNNs in implicit relation modeling and the non-i.i.d learning of GML in a unified framework. It first utilizes the distribution of instances learned by the state-of-the-art DNN-based model for ACD to encode similar semantic relations based on k-nearest neighborhood in the latent space. Complementarily, it uses a BERT-based model to extract the semantic relation concerning a given category, i.e., whether the two sentences are \emph{similar} or \emph{opposite}. Specifically, the category on-target is regarded as an auxiliary sentence, and a binary model is trained to predict the semantic orientation of two instances in response to the auxiliary sentence. The predicted relation can be either \emph{similar}, which means that both are relevant to the category or neither of them are, or \emph{opposite}, which means that one is relevant but the other is not. All the semantic relations are modeled as binary features in a factor graph. Finally, we conducted empirical experiments on real benchmark data through a comparative study to validate the efficacy of our proposed approach. Our extensive experiments demonstrated that our approach consistently outperforms existing methods, achieving state-of-the-art performance on all test datasets with considerable improvement margins. Overall, our results highlight the effectiveness of our proposed approach in addressing the ACD task.




Our main contributions are three-fold: 
\begin{enumerate}
	\item We  propose a novel supervised GML approach for the task of ACD, which can exploit the strength of DNN in semantic relation modeling to enable effective knowledge conveyance;
	
	\item We present a BERT-based binary model to predict the semantic relation concerning a given category, i.e. \emph{similar} or \emph{opposite}, between two sentences;
	
	\item We empirically validate the performance of the proposed approach on real benchmark data by a comparative study. Our extensive experiments have shown that it consistently achieves the state-of-the-art performance across all the test datasets and the improvement margins are considerable.
\end{enumerate}


%% file: related_work.tex
\section{Related work}

  The task of ACD has been extensively studied in the literature~\cite{hu2004mining,pontiki2016semeval,saeidi2016sentihood,jiang-etal-2019-challenge}. Unsupervised approaches for ACD typically aim to label each token in the text based on prior knowledge, such as the category's seed. For example, the spreading activation approach~\cite{schouten2017supervised} utilized co-occurrence association rule mining to facilitate activation value spreading in the co-occurrence graph. Similarly, a soft cosine similarity-based approach~\cite{ghadery2018unsupervised} leveraged the embedding space to compute the semantic similarity of a given word to the category seed. However, these approaches heavily rely on handcrafted features. The supervised approaches addressed ACD as a multi-class classification problem, whereas a set of classifiers, i.e., a classifier per category. The SVM-based models \cite{castellucci2014unitor,kiritchenko2014nrc} introduced a set of features, including n-grams, stemmed n-grams, character n-grams, non-contiguous n-grams, word cluster n-grams and lexicon features. However, these techniques cannot capture the semantic relations between tokens within the same sentence. With the rapid development of deep neural networks, various DNN models have been proposed to learn the specific-category representation in the corpus. A hybrid method \cite{zhou2015representation} introduced modeling the semantic relations based on domain-specific embeddings as hybrid features for a logistic regression classifier. A convolutional neural network-based features model\cite{toh2016nlangp} incorporated the automatically learnable features and the n-grams and POS tags to train one-vs-all linear classifiers. An LSTM equipped with a CNN layer  model \cite{xue2017mtna} addressed ACD and ACE simultaneously in a uniform framework. A Gated Recurrent Units (GRUs) equipped with a topic attention mechanism \cite{movahedi2019aspect} proposed to filter the aspect-irrelevant information away. 
	

Recently, fine-tuning of language models such as BERT \cite{devlin2018bert} has achieved state-of-the-art performance in various NLP tasks, including question answering \cite{lepikhin2020gshard}, machine translation \cite{khanuja2021muril}, and sentiment analysis \cite{liu2019roberta}. Several approaches have been proposed to jointly address sentiment identification and category detection, where a new class ('none') is introduced to indicate the absence of a given category in the sentence. Sun et al. (2019) proposed BERT-based fine-tuning models for the ACD task, treating it as a question answering task where the sentence is transformed into a question and the model is trained to predict the answer. \citet{wu2020context} proposed a context-guided BERT-based fine-tuning approach, which adopted a context-aware self-attention network to learn to distribute attention under different contexts. Despite the effectiveness of these approaches, learning implicit aspect-specific representations heavily relies on the availability of labeled examples, which may not be readily accessible in real-world scenarios. The GML paradigm was first proposed for the task of entity resolution \cite{hou2020gradual}, and has since been applied to aspect-based sentiment polarity detection \cite{wang2019gradual,ahmed-etal-2021-dnn}. These solutions are essentially unsupervised and do not rely on labeled training data for knowledge conveyance. However, the performance of unsupervised GML is often limited by inaccurate and insufficient knowledge conveyance. Unlikely, we introduce technical details to adopt supervised GML for ACD task.

Recently, fine-tuning of language models such as BERT \cite{devlin2018bert} has achieved state-of-the-art performance in various NLP tasks, including question answering \cite{lepikhin2020gshard}, machine translation \cite{khanuja2021muril}, and sentiment analysis \cite{liu2019roberta} and text classification \cite{abs-2306-07621,AhmedWAPSCL24} . Several approaches have been proposed to jointly address sentiment identification and category detection. In these approaches, a new class (i.e., None) is introduced to indicate the absence of a given category in the sentence. \citet{sun2019utilizing} proposed BERT-based fine-tuning models for the ACD task, treating it as a question answering task where the sentence is transformed into a question and the model is trained to predict the answer. \citet{wu2020context} proposed a context-guided BERT-based fine-tuning approach, which adopted a context-aware self-attention network to learn to distribute attention under different contexts.
Despite the effectiveness of these approaches, learning implicit aspect-specific representations heavily relies on the availability of massive labeled examples, which may not be readily accessible in real-world scenarios. GML paradigm was first proposed for the task of entity resolution \cite{hou2020gradual}, and has since been applied to aspect-based sentiment analysis detection \cite{wang2019gradual,ahmed-etal-2021-dnn}. These solutions are essentially unsupervised and do not rely on labeled training data for knowledge conveyance. However, the performance of unsupervised GML is often limited by inaccurate and insufficient knowledge conveyance. Therefore, in this work, we introduce technical details to adopt supervised GML for the ACD task.


%% file: background.tex
\section{Preliminaries}\label{sec:background}


%
In this section, we first define the ACD task, and provide an overview of the GML paradigm. 


\subsection{Task Statement}

  The task of ACD is formally defined as follows:
\begin{definition} 	
[{\bf Aspect Category Detection}]  
Let $S=\{s_1, s_2, \ldots, s_n\}$ be a corpus of sentence reviews, where each $s_i$ is annotated with at least one aspect category from a predefined set $C=\{c_1, c_2, \ldots, c_m\}$. The goal of the ACD task is to predict the label of each pair ($s_i, c_j$) as $y=\{0, 1\}$, where $y=1$ indicates relevance and $y=0$ indicates irrelevance. It is noteworthy that in some datasets (e.g., SentiHood \cite{jiang-etal-2019-challenge}), a category may be specified by a pair, ${(t,c) : (t \in T, c \in C)}$, where $T$ is a predefined set of targets (e.g., LOCATION\#1, LOCATION\#2) that represent the names of neighborhoods.
\end{definition}

\subsection{GML Paradigm Overview}

\begin{figure}
	\centering
	\includegraphics[scale=.63]{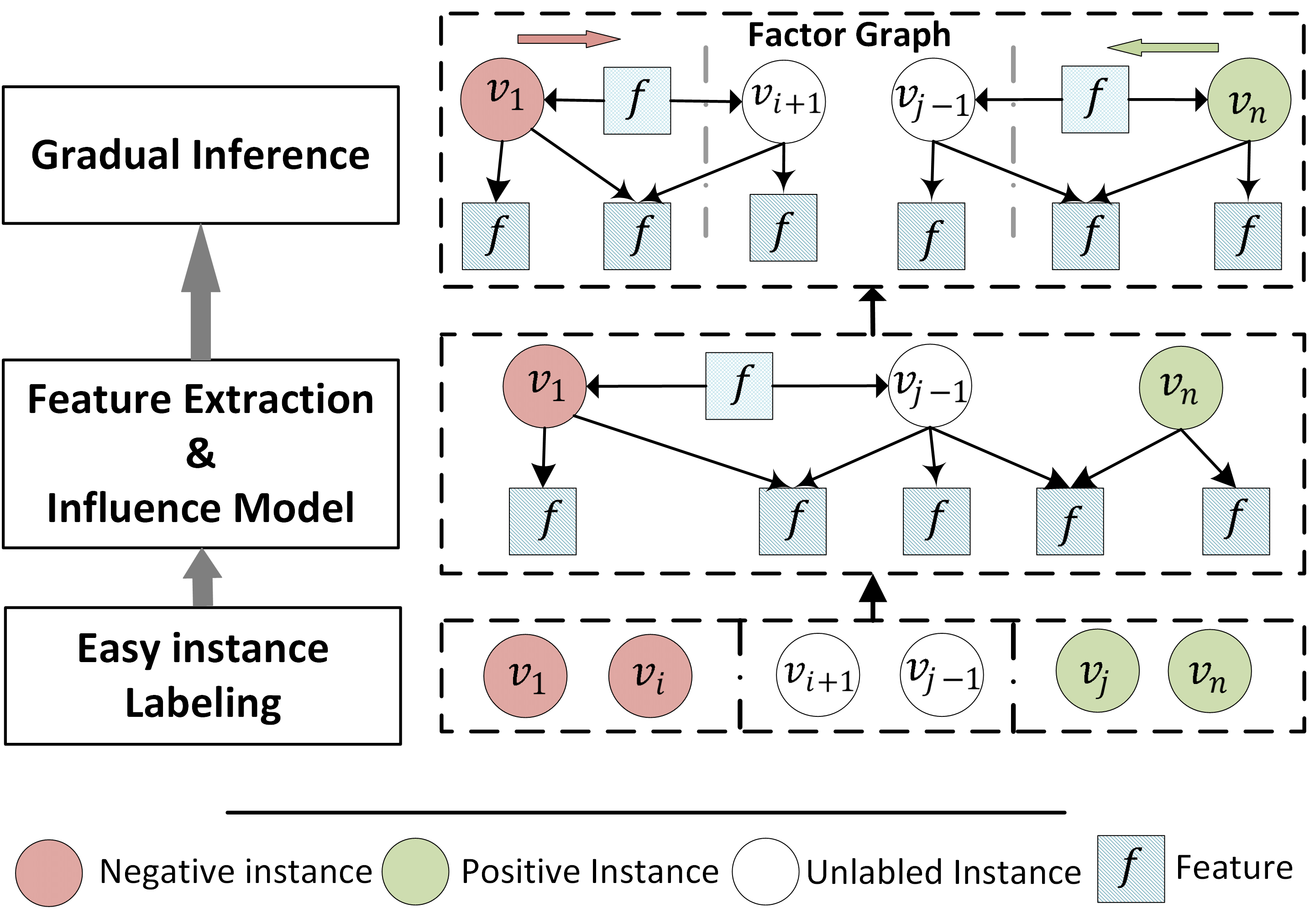}	
	\caption{ An illustrative example of the GML paradigm. In brief, GML consists three steps. Firstly, easy instance labeling relies on simple rules or unsupervised learning to automatically label some instances in the task. Secondly, feature extraction and influence modeling extract common features from labeled and unlabeled instances and model their influence on labels. Lastly, gradual inference labels instances with increasing difficulty using iterative inference, choosing the instance with the highest degree of evidential certainty at each iteration until all instances are labeled. }
	\label{fig:gml}
\end{figure}

As shown in Figure~\ref{fig:gml}, GML consists of the following three steps:

\subsubsection{Easy Instance Labeling}

Accurately labeling all instances in a classification task can be challenging, especially without good-coverage training examples. However, the task can become easier if only a subset of easy instances needs to be automatically labeled. In real-world scenarios, easy instance labeling can be performed using simple user-specified rules or existing unsupervised learning techniques. GML begins with observations provided by the labels of these easy instances. Therefore, high accuracy in automatic machine labeling of easy instances is critical for GML's performance on a given task. In our GML solution for ACD, we use supervised learning, and we consider the training examples as easy instances.

\subsubsection{Feature Extraction and Influence Modeling}
Feature serves as the medium for knowledge conveyance. This step extracts the common features shared by labeled and unlabeled instances. To facilitate effective knowledge conveyance, it is desirable that a wide variety of features are extracted to capture as much information as possible. For each extracted feature, this step also needs to model its influence over the labels of its relevant instances.


\subsubsection{Gradual Inference} 

The gradual learning process involves labeling instances in a task gradually, starting with the easiest and progressing to the more difficult ones. As gradual learning does not satisfy the i.i.d assumption, it is approached from the perspective of evidential certainty. This is accomplished by conducting gradual learning over a factor graph that includes both labeled and unlabeled instances, along with their common features. Iterative inference is employed, whereby the unlabeled instance with the highest degree of evidential certainty is selected and labeled at each iteration. This iterative process is repeated until all the instances in a task are labeled.
 

To implement scalable gradual inference, the unlabeled variables are initially sorted based on their evidential support. The top $m$ variables are then selected as candidates for probability inference. To reduce the frequency of factor graph inference, an efficient algorithm is employed to approximate entropy estimation on the $m$ candidates and select the top $k$ most promising variables for factor graph inference. The probabilities of the selected $k$ variables are then inferred in the subgraphs of G.

%% file: approach.tex
\section{Supervised GML for ACD}
\begin{figure*}[t]
	\centering
	\includegraphics[scale=0.5]{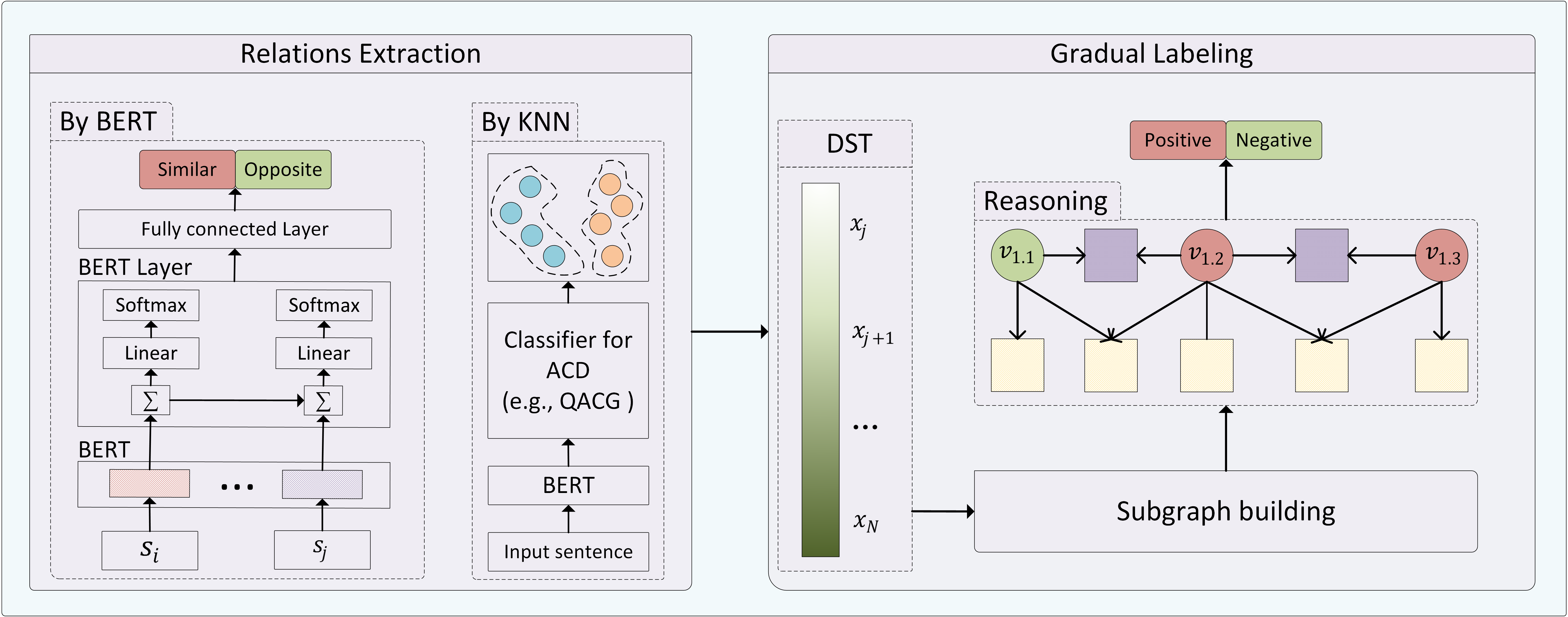}	
	\caption{An example of the proposed solution, which consists of two main steps. Firstly, it identifies common semantic features among instances by modeling two types of relations: BERT-based and KNN-based relations. These semantic connections between instances provide a high-level description of whether two instances discuss the same aspect. Secondly, the instances are gradually labeled based on their level of difficulty. The gradual labeling process begins with the easiest instances and progresses to the more challenging ones. The information gained from labeling the easier instances is then used to guide the labeling of the harder ones. This way, the labels assigned to the easier instances serve as a reference for labeling the more difficult ones, resulting in a more efficient and accurate labeling process.}
	\label{fig:framework}
\end{figure*}

In this section, we first present the approach to extract semantic relations between a pair of sentence reviews using DNN models. We then describe how to incorporate these relations into the gradual learning process in a factor graph. An example of the proposed solution is depicted in Figure \ref{fig:framework}.

\subsection{Semantic Relation Extraction}


The proposed approach utilizes binary relations between a pair of instances to facilitate knowledge transfer. As a sentence may pertain to multiple categories simultaneously, we extract binary relations for each category independently. For a given category $c$, a pair of instances $s_i$ and $s_j$ are considered \emph{similar} with respect to $c$ if they are both relevant or irrelevant to $c$. Conversely, if one instance is relevant to $c$ while the other is not, they are considered \emph{opposite} with respect to $c$. We first use a state-of-the-art DNN-based model for ACD that learns category-specific vector representations of sentences, to extract similar binary relations based on k-nearest neighborhood.  additionally, we train a special Bert-based binary model using labeled examples to detect category-specific relations, which can be \emph{similar} or \emph{opposite}, between two  arbitrary instances.
\subsubsection{By Category Classification DNN}

Contextual PLMs, such as BERT, are able to learn contextual representations of words, which can capture the meaning of a word based on its surrounding context. This enables these models to map features that occur in similar contexts into close points in an embedding space \cite{ahmed2020constructing}.
State-of-the-art DNN solutions for ACD often utilize PLMs, to initialize feature vectors for sentences or phrases. These feature vectors can then be fine-tuned during training to learn category-specific representations that are optimized for the specific ACD task at hand. The goal of fine-tuning the pre-trained feature vectors is to ensure that instances relevant to the same category are located very close in the latent space. This is achieved by adjusting the weights of the neural network during training, so that the feature vectors are optimized to capture the underlying structure and relationships of the data \cite{sun2019utilizing}. 

\begin{figure}[t]
	\centering
	\includegraphics[scale=.55]{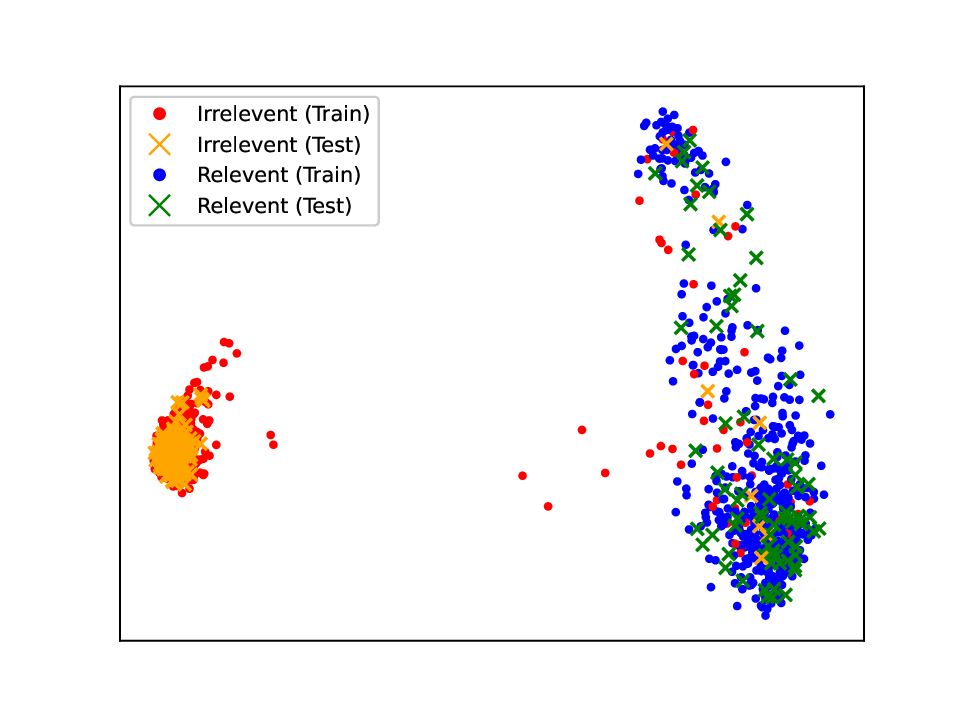}	
	\caption{The distribution of category \textit{Menu}'s instances based on final layer of QACG-BERT. }
	\label{fig:knn}
\end{figure}

Figure~\ref{fig:knn} presents an illustrative example that visualizes the distribution of instances related to the category of \textit{Menu} by the QACG-BERT model \cite{wu2020context}. It can be observed that the instances, whether from the labeled training set or the unlabeled test set, are mostly mapped based on their relevance to \textit{Menu}. Motivated by this, we leverage the category DNN classifier to extract semantic similar relations. For an unlabeled instance in a workload, we generate its \textit{k}-nearest neighbors instances from the training set based semantic distance. In the implementation, we set a threshold (e.g., 0.9 in our experiments) to filter out the uncertain instances. The resulting similar relations indicate that either both instances in a pair are relevant or neither of them are. Specifically, our implementation leverages the state-of-the-art QACG-BERT model \cite{wu2020context} for similar relation extraction. However, other DNN models can be easily leveraged in the same way.
	                                                              
\subsubsection{By Bert-based Binary Models.}

 The ultimate objective is to estimate the relationship between two arbitrary sentences related to a given category, represented as a query in BERT's notation \cite{devlin2018bert}. BERT-based models for aspect-based sentiment analysis (ABSA) mostly rely on the ``[CLS]'' token to represent the sentence representation. However, a \citet{jawahar-etal-2019-bert} have studied on the language structure learned by BERT and demonstrated that the middle layers can capture syntactic features, while semantic features can be extracted from the top layers. Since ACD involves learning semantic connections that need to be captured at the higher layers of the network, we leverage the top layers of BERT to model semantic relations.
 


Inspired by the Feature Pyramid Networks (FPNs) \cite{lin2017feature}, we establish a hierarchical connection between layers, as illustrated in Figure \ref{fig:framework}. We approach the semantic relation modeling of each category as a binary classification problem, where each category on-target $c$ serves as an auxiliary sentence. To form the input, we concatenate the pair of sentences using the separator "[SEP]" to create $x$. The hidden layers of BERT are as follows, given such input:
\begin{equation}\label{eq:bert}
Z = BERT(x,c),
\end{equation}
where $Z$ denotes the hidden layers. We exploit the top four layers of $Z$ and  perform predictions for each layer separately. Particularly, we add one more BERT layer that takes the previous and the current layers as input. The intuition behind this architecture is that the deeper layers contain the most semantic information in response to the category on-target \cite{lin2017feature}. Therefore, establishing an elaboration between layers can capture the deep semantic representations at different levels \cite{karimi2020improving}. The output of the added layer is aggregated with the previous layer. Meanwhile, a prediction is performed to estimate the final loss, which is the sum of the entropy loss of the four top layers.  
For each category, we train a binary model (shown in Figure \ref{fig:framework}) using labeled examples. Once the model is trained, we randomly extract $k_b$ relations from both the training and test sets for each instance in the test set. Our experiments have demonstrated that the performance of supervised GML is highly robust with respect to the value of $k_b$, provided that it falls within a reasonable range of $3\le k_b\le 11$.

\begin{figure*}
	\centering
	\includegraphics[width=1\linewidth]{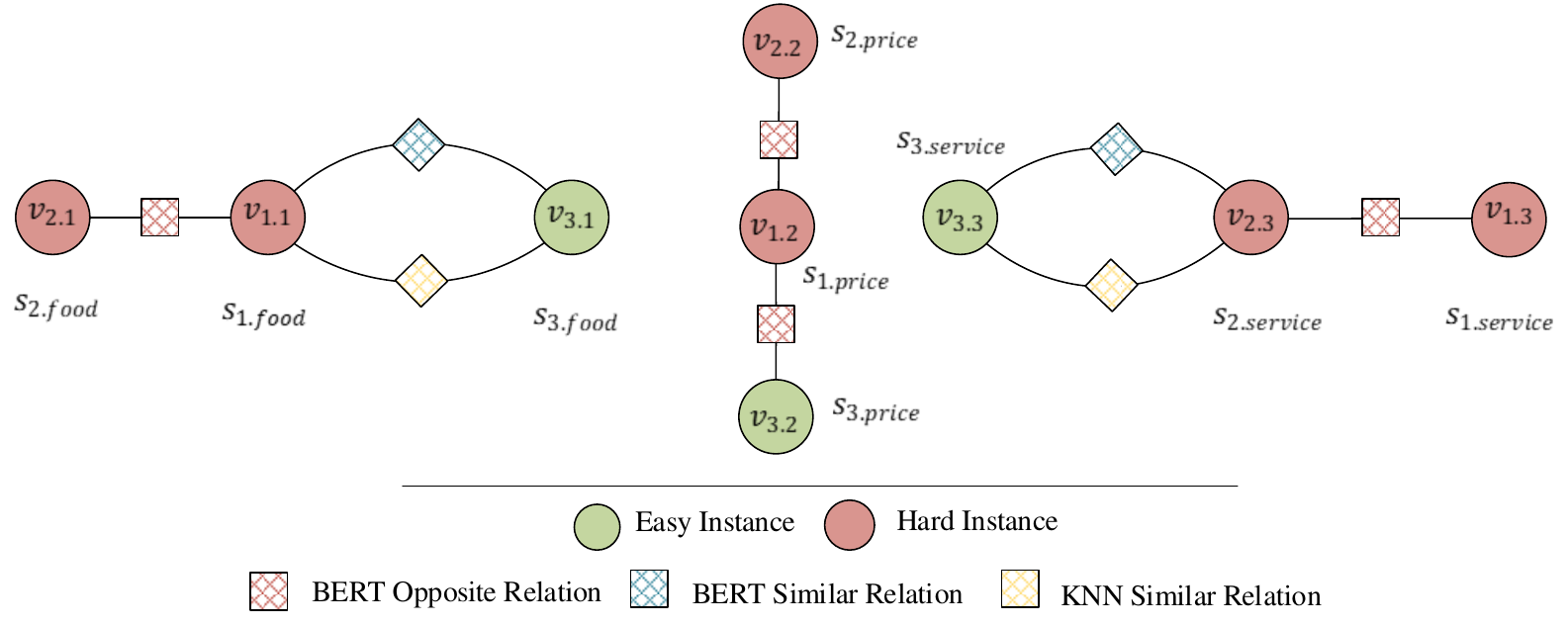}
	\caption{An example of gradual inference of the running example.}
	\label{fig:Gradual}
	
\end{figure*}

\subsection{Factor Modeling of Binary Relations}



Based on the paradigm of GML, each sentence is represented by $M$ variables in the factor graph $G$, where $M$ is the number of predefined categories. Each variable is assigned a value representing its relevance, with 1 indicating relevance and 0 indicating irrelevance. Note that the values of evidence variables, once inferred, remain unchanged during the inference process, while the values of inference variables need to be gradually inferred based on G. The corresponding factor graph of the running example is presented in Figure \ref{fig:Gradual}.

In the supervised setting, the initial easy instances are the labeled examples in the training set. The labels of the instances in the test set are then gradually inferred by conveying knowledge through semantic relations. Formally, the factor of a semantic relation $f_r$ is defined as:
  \begin{equation}
  \label{eq:binary factor}
  \varphi_{f_r}(v_i, v_j) = 
  \left \{
  \begin{array}{ll}
  e^{w_{f_r}}     &      if \ v_i = v_j; \\
  1    &   otherwise;  
  \end{array} 
  \right.
  \end{equation}
where $v_i$ and $v_j$ denote the two variables sharing the relational feature $f_r$, and $w_{f_r}$ denotes the weight of $f_r$.

Formally, given a factor graph for ACD, $\textrm{G}$, GML defines the probability distribution over its variables $V$ as follows:
\begin{equation}
\label{eq:fg_model}
P_w(V)=\frac{1}{Z_w}\prod_{v \in V}
\prod_{f_r \in F_r}\phi_{f_r}(v_i, v_j),
\end{equation}
where  $F_r$ denotes the set of the modeled relations, $\phi_{f_r}(v_i, v_j)$ denotes the factor associated with the relational feature $f_r$, and $Z$ is a partition function, i.e. normalization constant.

Note that the weight of a similar relational factor is positive and the weight of an opposite relational factor is negative. Both weights needs to be learned in the inference process based on evidential observations. To effectively learn the factor weights without access to the true labels of unlabeled variables, $V_I$, GML minimizes the negative log marginal likelihood given the observed labels of labeled variables, $\Lambda$, as follows:
\begin{equation} \label{eq:weight-learning}
\hat w  = arg \min \limits_{w} -log \sum_{V_I} P_w(\Lambda, V_I).
\end{equation}

%% file: experiment.tex
\section{Empirical  Evaluation}
\begin{table}[t]
	\centering
		\adjustbox{max width=\linewidth}{
	\begin{tabular}{l|l|c|c|c}
		\hline
		Dataset&Train &Test& \makecell[c]{\# Category}& \makecell[c]{MultiCat (\%) }
		\\
		\hline
		
		SemEval 14 &3041&800&4&23.65 \%\\
		SemEval 16 &2000&676&12&26.18\%\\
		MAMS &3713&901&8&100\%\\
		SentiHood &3724&1491&4&31.0 \%\\
		\hline
	\end{tabular}	}
	\caption{The statistics of test datasets. The column of (MultiCat) Multiple Category reports the percentage of sentences labeled with more than one category.}
	\label{tab:datasets}
\end{table}

\subsection{Datasets}

\begin{table*}[]
	\centering
	
	\adjustbox{max width=\linewidth}{
		\begin{tabular}{l|l|c|c|c|c|c|c|c|c|c}
			\hline
			
			\multirow{2}{*}{Initializer }&\multirow{2}{*}{Model }& \multicolumn{3}{c|}{SemEval 2014}&\multicolumn{3}{c|}{SemEval 2016} &\multicolumn{3}{c}{MAMS}\\\cline{3-11}
			
			&&P&R & F1&P&R & F1  &P&R & F1 \\
			\hline
			%
			\multirow{4}{*}{Bert-based }
			&BERT-pair-NLI-B  &93.15	&90.24&	91.67&	88.79&	84.92&	86.71&93.89	&91.88&	92.8\\
			&BERT-pair-NLI-M&93.57&	90.83&	92.18&86.03&	88.35&	87.18&92.88&	93.36&	93.11\\
			&CG-BERT &93.12&	90.17&	91.62&	84.55&	90.04&	87.01&92.36	&92.44&	92.4\\
			&QACG-BERT &94.27&	90.12&	92.14&	86.52&	88.50&	87.48& 90.85&	93.74&	92.1\\
			
			\hline
			\multirow{2}{*}{GML-based }&GML (w/o KNN) &94.59	&92.35	&93.46	&88.64&	89.72&	89.17&	92.92&	93.61	&93.26\\
			
			& GML & 	\textbf{94.63}&	\textbf{92.59}&\textbf{	93.60}&	\textbf{88.69}&\textbf{	90.40}&	\textbf{89.54}&\textbf{	93.30}&	\textbf{93.82}&	\textbf{93.55}	\\
			\hline
			
	\end{tabular}}
	\caption{The evaluation results on SemEval and MAMS. P, R, and F1 represent Precision, Recall and F1 scores, respectively.   The best scores are highlighted in {\bf bold}.}
	\label{tab:results_absa}
\end{table*}

\begin{table}
	\centering
	\adjustbox{max width=\linewidth}{
		\begin{tabular}{l|l|c|c}
			\hline
			&Model& Strict Accuracy & F$_1$ score\\
			\hline
			\multirow{4}{*}{Bert-based }
			&BERT-single &73.7&81.0\\
			&BERT-pair-NLI-M  &79.4&87.0\\
			&BERT-pair-NLI-B &79.8&87.9\\
			&CG-BERT  &80.1&88.84\\
			&QACG-BERT  &80.9 &90.63\\
			
			\hline
			\multirow{2}{*}{GML-based }& GML (w/o KNN) &81.31&93.48\\
			&GML &\textbf{81.38}&\textbf{93.69}\\
			\hline
			
	\end{tabular}}
	\caption{The evaluation results on SentiHood. The best scores are highlighted in {\bf bold}. }
	\label{tab:results_tabsa}
	
\end{table}


	
	We use the benchmark datasets of SemEval 2014 Task 4\footnote{SemEval-2014 Task 4: \url{https://alt.qcri.org/semeval2014/task4/}.},  SemEval 2016 task 5\footnote{SemEval-2016 Task 5: \url{https://alt.qcri.org/semeval2016/task5/}.} and MAMS\footnote{MAMS: \url{https://github.com/siat-nlp/MAMS-for-ABSA}.}  \cite{pontiki2014semeval,pontiki2016semeval,jiang-etal-2019-challenge}, which contain the restaurant reviews collected from Citysearch New York corpus, as well as the SentiHood dataset\footnote{Sentihood dataset: \url{https://github.com/uclnlp/jack/tree/master}.}
\cite{saeidi2016sentihood}, which was built from the Question Answering Yahoo corpus with location names of London and UK. The SentiHood dataset consists of reviews that evaluate at least one aspect category $c$ in response to the target $t$ (e.g., LOCATION\#1) that represents a location name of the neighborhoods. The statistics of the test datasets are detailed in Table \ref{tab:datasets}. Each dataset is partitioned into train, validation and test sets as in the benchmark project or its original presentation paper.

\subsection{Experimental Setup}

\begin{table*}
	\centering
	
	\begin{tabular}{c|c|c|c|c|c|c|c|c}
		\hline
		\multirow{2}{*}{Proportion} &\multicolumn{2}{c|}{SemEval 2014}&\multicolumn{2}{c|}{SemEval 2016}&\multicolumn{2}{c|}{MAMS}&\multicolumn{2}{c}{SentiHood}\\ \cline{2-9}
		&QACG&GML&QACG&GML&QACG&GML&QACG&GML\\
		\hline
		20\%&89.13&92.24&82.09&84.13&90.0&90.94&87.64&89.21 \\
		30\%&91.31&93.08&85.33&87.34&89.93&91.96&88.71& 91.69\\
		50\%&92.05&93.34&86.32&88.59&91.93&92.4&89.63&92.27 \\
		100\%&92.14&93.46&87.48&89.17&92.1&93.26&90.63&93.48 \\
		\hline
	\end{tabular}
	\caption{The performance of GML compared to  QACG-BERT trained on different sufficiency levels of training data in terms of Macro-F1.}
	\label{tab:propotion}		
\end{table*}

   \begin{figure*}
	\centering
	\includegraphics[scale=.6]{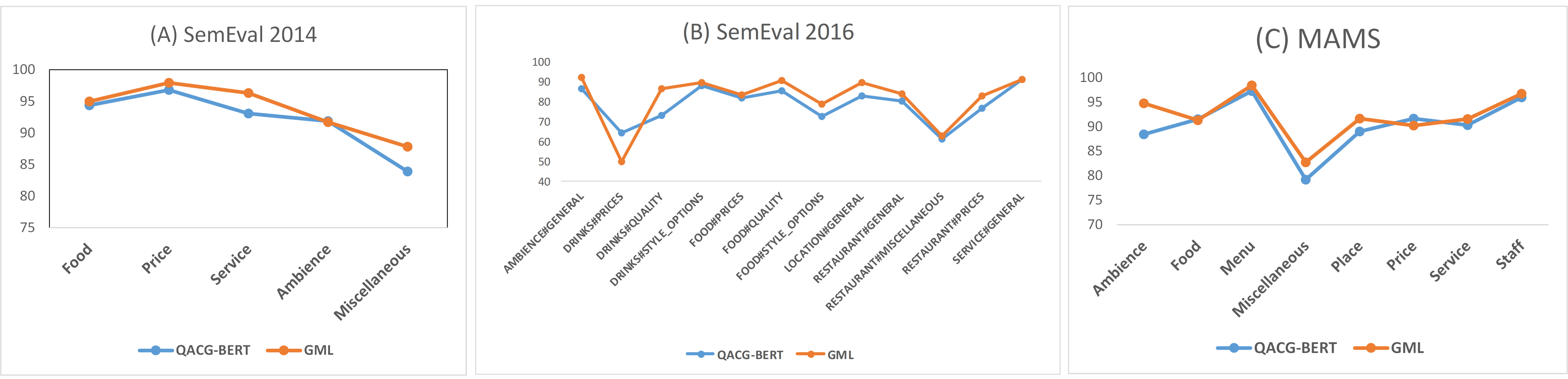}	
	\caption{The performance of GML on each category of SemEval and MAMS compared with QACG-BERT in terms of Macro-F1.} 
	\label{fig:absa_task}
\end{figure*}

 Because the Bert-based models for ACD have been empirically shown to perform better than earlier proposals, we compare the proposed approach (denoted by GML) with these state-of-the-art BERT-based models, which include 

\begin{itemize}
	
	
	
	\item BERT-pair-NLI-M \cite{sun2019utilizing}. It is a  BERT-based fine-tuning model, which converts ACD to a sentence-pair classification task. It was trained to learn the specific-category representation as a pseudo-sentence natural language inference.
	
	\item 	BERT-pair-NLI-B \cite{sun2019utilizing}. It is a variant of BERT-pair-NLI that uses the category as an auxiliary sentence and then addresses ACD as a binary classification problem $(label \in \{yes, no\})$ to obtain the probability distribution.
	\item CG-BERT \cite{wu2020context}. It is a context-guided BERT-based fine-tuning model, which adopts a context-aware self-attention network. It incorporates the context into the calculation of attention weights to deeply learn the  hidden representations.
	
	\item 	QACG-BERT  \cite{wu2020context}. QACG-BERT is an improved variant of CG-BERT model that introduces learning quasi-attention weights in a compositional manner to enable subtractive attention lacking in softmax-attention.
\end{itemize}

     We used the QACG-BERT model \cite{wu2020context} to extract similar semantic relations. In the training set, for each category, we generate 10 semantic relations (5 for similar and 5 for opposite) per each labeled example. The resulting labeled pairs serve as the training data for the Bert-based binary model. For gradual learning, we randomly select 5 instances for each unlabeled instance from both the training and test sets, and use the trained binary model to label their semantic relations. We set the nearest neighbors $k$ to 3.

		Following the settings of the original BERT-base model \cite{devlin2018bert}, the bert-based model for general semantic relation detection consists of 12 heads and 12 layers with the hidden layer size of 768. As a result, the total number of its parameters is 138M. When fine-tuning, we keep the dropout probability at 0.1 and set the number of epochs to 3. The initial learning rate is $5e^{-5}$ for all layers with a batch size of 32. We initialize the model with the post-trained BERT, which has been trained using uncased version of BERT-BASE \cite{karimi2021adversarial}. 
All the reported results are averages over 3 runs. Our experiments show that the performance of evaluated models only fluctuates marginally in different runs.

  For the the evaluation on SemEval and MAMS, as presented in the task descriptions \cite{pontiki2014semeval,pontiki2016semeval,jiang-etal-2019-challenge}, we compare the performance measured by Precision, Recall, and F1 scores. For the evaluation on the SentiHood dataset, for a fair comparison,  we follow the experimental settings of previous work \cite{saeidi2016sentihood,ma2018targeted,sun2019utilizing,wu2020context}. Only four aspect categories, i.e., most frequently appear in the corpus,  are considered, i.e., general, price, transit-location and safety. Similarly, we report the results of strict accuracy metric and Macro-F1 score. Strict accuracy requires the model to correctly detect all aspect categories to a given target, while Macro-F1 is the harmonic mean of the Macro-precision and Macro-recall of all targets.

\subsection{Evaluation Results}

 The detailed evaluation results on SemEval and MAMS are reported in Table~\ref{tab:results_absa}. It can be observed that GML consistently achieves the state-of-the-art performance across all datasets in terms of Precision, Recall and F1 scores. Specifically, in terms of F1, GML outperforms the previous state-of-the-art by around 1.5\% on SemEval 2014 and MAMS, the improvement margin is even larger at 2\% on SemEval 2016. Regarding the well-known challenge of ACD, these improvement margins are considerable. We also report the performance of GML on each category compared to the QACG-BERT \cite{wu2020context} in Figure \ref{fig:absa_task}. It can be observed that GML performs better than QACG-BERT on most categories with only a few exceptions.

\begin{figure}
	\centering
	\includegraphics[scale=.6]{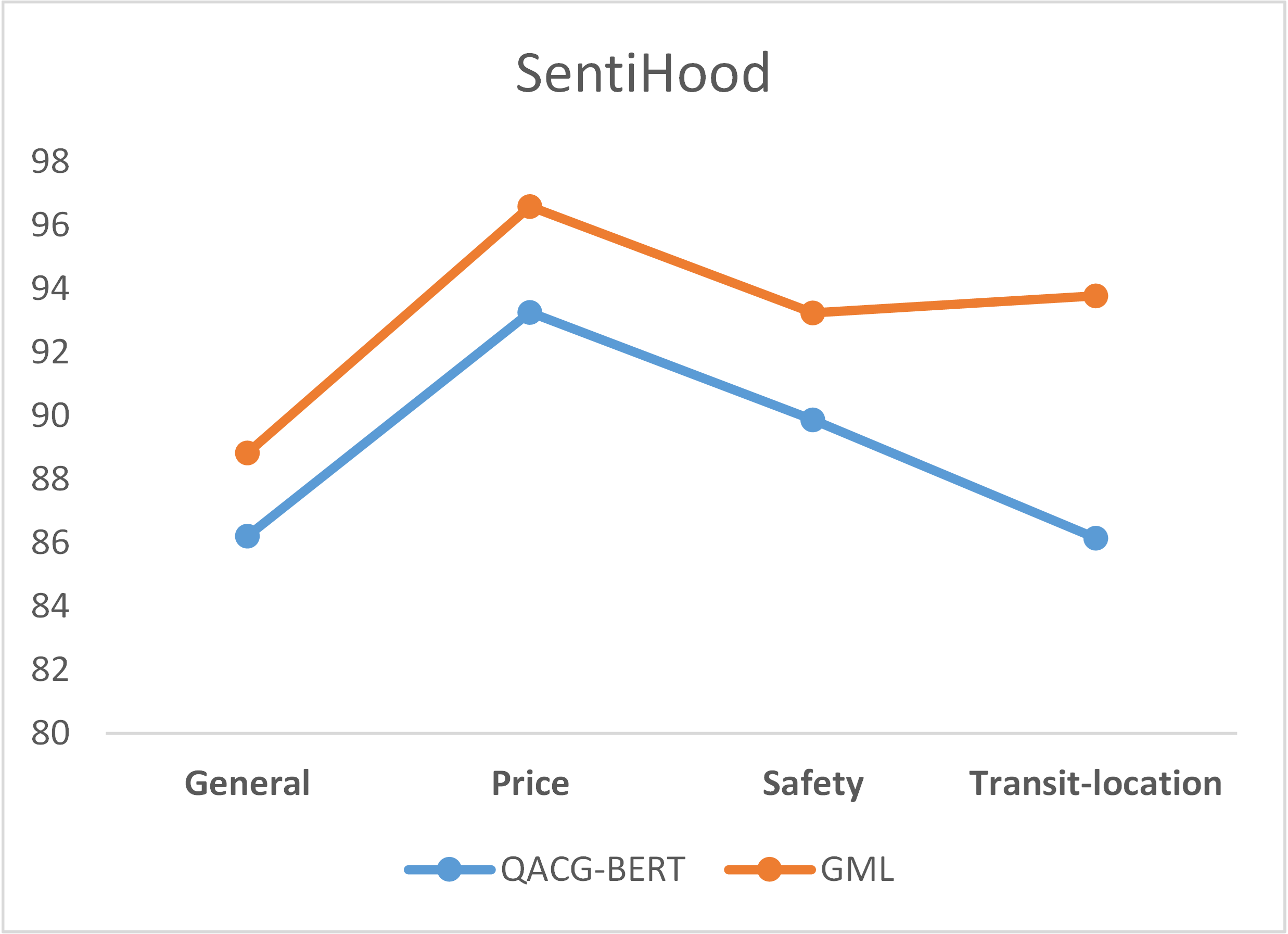}	
	\caption{The performance of GML compared to QACG-BERT on each category of SentiHood in terms of Macro-F1.}
	\label{fig:tabsa_task}
\end{figure}
  The detailed evaluation results on SentiHood have been presented in Table~\ref{tab:results_tabsa}. It can be observed that the proposed solution consistently achieves the state-of-the-art performance in terms of both Strict Accuracy and Macro-F1 at 81.38\% and 93.69\% respectively.
It is noteworthy that GML improves the state-of-the-art by more than 2\% in terms of F1. Moreover, Figure \ref{fig:tabsa_task} reports the performance of GML on each category compared to the QACG-BERT. It can be observed that GML outperforms QACG-BERT by comfortable margins on all categories. 

  We have also conducted an ablation test on the GML solution. Since the extracted KNN similar relations are usually sparse, semantic relation extraction by the Bert-based binary model is essential to the efficacy of the proposed solution. Therefore, we report the performance of GML without KNN semantic relations in Table~\ref{tab:results_absa} and \ref{tab:results_tabsa}. It can be observed that KNN semantic relations are helpful to the performance of GML. 

\subsubsection{Evaluation with different sufficiency levels of training data}  

  We compare GML with QACD-BERT in various settings with different sufficiency levels of labeled examples. We set the sufficiency levels at 20\%, 30\%, 50\% and 100\% of the original training set. The detailed results are presented in Table \ref{tab:propotion}. It can be observed that with only 30 \% of the training set, GML can achieve a competitive performance across all datasets compared to QACD-BERT with the whole training set. Our experimental results clearly demonstrate that a carefully-designed collaboration between GML and DNN can effectively perform better than DNN itself.

 \begin{table}
	\centering
	\adjustbox{max width=\linewidth}{
		\begin{tabular}{c|c|c|c|c}
			\hline
			\makecell[c]{\#  Relation}& \makecell[c]{SemEval 2014}&\makecell[c]{SemEval  2016}&MAMS&SentiHood\\
			\hline
			3&	93.43&	89.3&	93.38&93.39\\
			5&	93.46&	89.17&	93.26&93.85\\
			7&	93.47&	89.91&	93.46&93.63\\
			9&	93.32&	90.19&	93.63&93.7\\
			11&	93.45&	89.9&	93.49&93.45\\
			
			\hline
	\end{tabular}}
	\caption{Sensitivity evaluation of GML w.r.t the number of semantic relations extracted by the Bert-based binary model, with performance in terms of Macro-F1.}
	\label{tab:n_rel}		
\end{table}




\subsubsection{Sensitivity Evaluation}

We have also evaluated the performance sensitivity of GML w.r.t the number of semantic relations extracted by the Bert-based binary model. We vary the value of $k_b$ from 3 to 11. As shown in Table \ref{tab:n_rel}, the performance of GML is very robust. These experimental results bode well for its applicability in real scenarios.